\documentclass{article}
\usepackage{spconf,amsmath,graphicx}

\usepackage{amssymb}
\usepackage[ruled,linesnumbered]{algorithm2e}
\usepackage{booktabs}
\usepackage{float}
\usepackage{bigstrut}
\usepackage{multirow}
\usepackage{balance}
\usepackage[hidelinks=true,bookmarks=false]{hyperref}
\usepackage{lineno}
\usepackage{amsopn}
\usepackage{comment}
\usepackage{color}
\usepackage{algorithmic}
\usepackage{tabularx}
\usepackage{mathrsfs}
\usepackage{makecell}
\usepackage{url,subfigure,tabulary}
\usepackage{color}
\usepackage{algorithmic}
\usepackage{enumitem}

\newcommand{\eg}{\textit{e.g.}}

\newcolumntype{L}[1]{>{\raggedright\arraybackslash}p{#1}}
\newcolumntype{C}[1]{>{\centering\arraybackslash}p{#1}}
\newcolumntype{R}[1]{>{\raggedleft\arraybackslash}p{#1}}

\usepackage{pifont}
\newcommand{\cmark}{\ding{51}}%

\title{Co-Teaching: An Ark to Unsupervised Stereo Matching}

\name{Hengli Wang$^{\star}$ \qquad Rui Fan$^{\dagger}$ \qquad Ming Liu$^{\star}$
}
  
\address{$^{\star}$ Hong Kong Unviersity of Science and Technology, Hong Kong SAR, China\\
      $^{\dagger}$ Tongji University, Shanghai 201804, China\\
      \normalsize\texttt{hwangdf@connect.ust.hk, rui.fan@ieee.org, eelium@ust.hk}}

\begin{document}
%
\maketitle
\begin{abstract}
  Stereo matching is a key component of autonomous driving perception. Recent unsupervised stereo matching approaches have received adequate attention due to their advantage of not requiring disparity ground truth. These approaches, however, perform poorly near occlusions. To overcome this drawback, in this paper, we propose CoT-Stereo, a novel unsupervised stereo matching approach. Specifically, we adopt a co-teaching framework where two networks interactively teach each other about the occlusions in an unsupervised fashion, which greatly improves the robustness of unsupervised stereo matching. Extensive experiments on the KITTI Stereo benchmarks demonstrate the superior performance of CoT-Stereo over all other state-of-the-art unsupervised stereo matching approaches in terms of both accuracy and speed. Our project webpage is \url{https://sites.google.com/view/cot-stereo}.
\end{abstract}


\section{Introduction}
\label{sec.introduction}
Stereo matching is a fundamental problem in computer vision and robotics. This important technique has been widely employed in many tasks, such as robot vision \cite{wang2019self,wang2020applying,wang2021dynamic} and autonomous driving \cite{fan2020sne,fan2021learning}. The goal of stereo matching is to estimate dense correspondences between a pair of stereo images and further generate a dense disparity image \cite{fan2018road}.

Traditional and data-driven approaches are two major types of stereo matching algorithms \cite{fan2018road,wang2021pvstereo}. Traditional algorithms formulate stereo matching as either a local block matching problem or a global energy minimization problem \cite{fan2018road}. Data-driven approaches \cite{psmnet,gwcnet,cheng2020hierarchical} utilize convolutional neural networks (CNNs) to extract informative visual features and create a 3D cost volume, by analyzing which a dense disparity image can be estimated. Among data-driven approaches, PSMNet \cite{psmnet} adopts 3D CNNs to regularize cost volumes for disparity estimation, while GwcNet \cite{gwcnet} further utilizes group-wise correlation to provide efficient representations for visual feature similarity measurement. Meanwhile, LEAStereo \cite{cheng2020hierarchical} uses a neural architecture search framework to search an effective and efficient network for stereo matching. However, such supervised stereo matching approaches typically require a large amount of training data with disparity ground truth, often making them difficult to apply in practice.

\begin{figure}[t]
  \centering
  \includegraphics[width=0.99\linewidth]{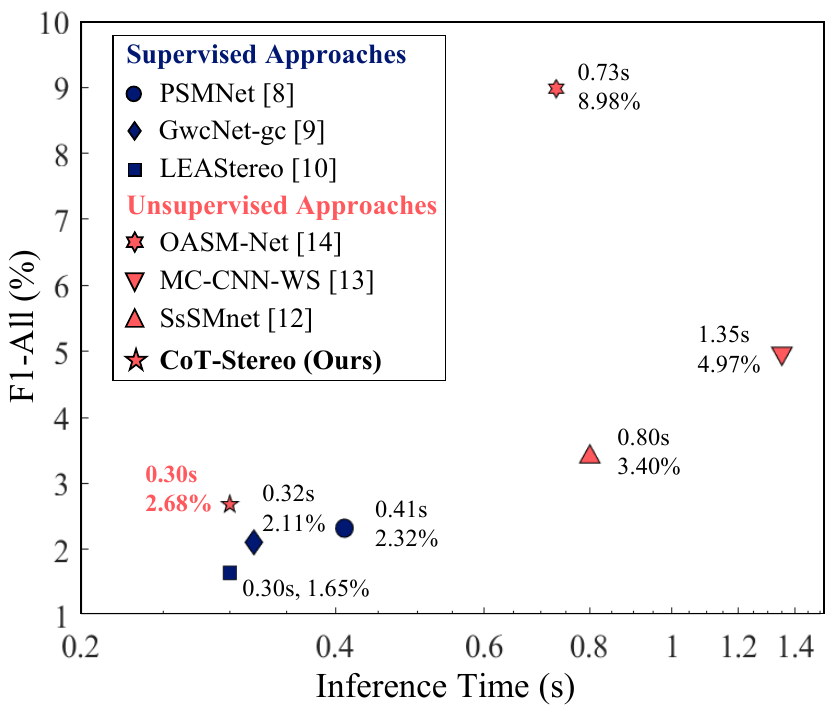}
  \caption{Evaluation results on the KITTI Stereo 2015 benchmark \cite{kitti15}, where ``F1-All'' denotes the percentage of erroneous pixels measured over all regions. Our CoT-Stereo outperforms all other state-of-the-art unsupervised stereo matching approaches in terms of both accuracy and speed.}
  \label{fig.time_error}
\end{figure}

\begin{figure*}[t]
  \centering
  \includegraphics[width=0.95\textwidth]{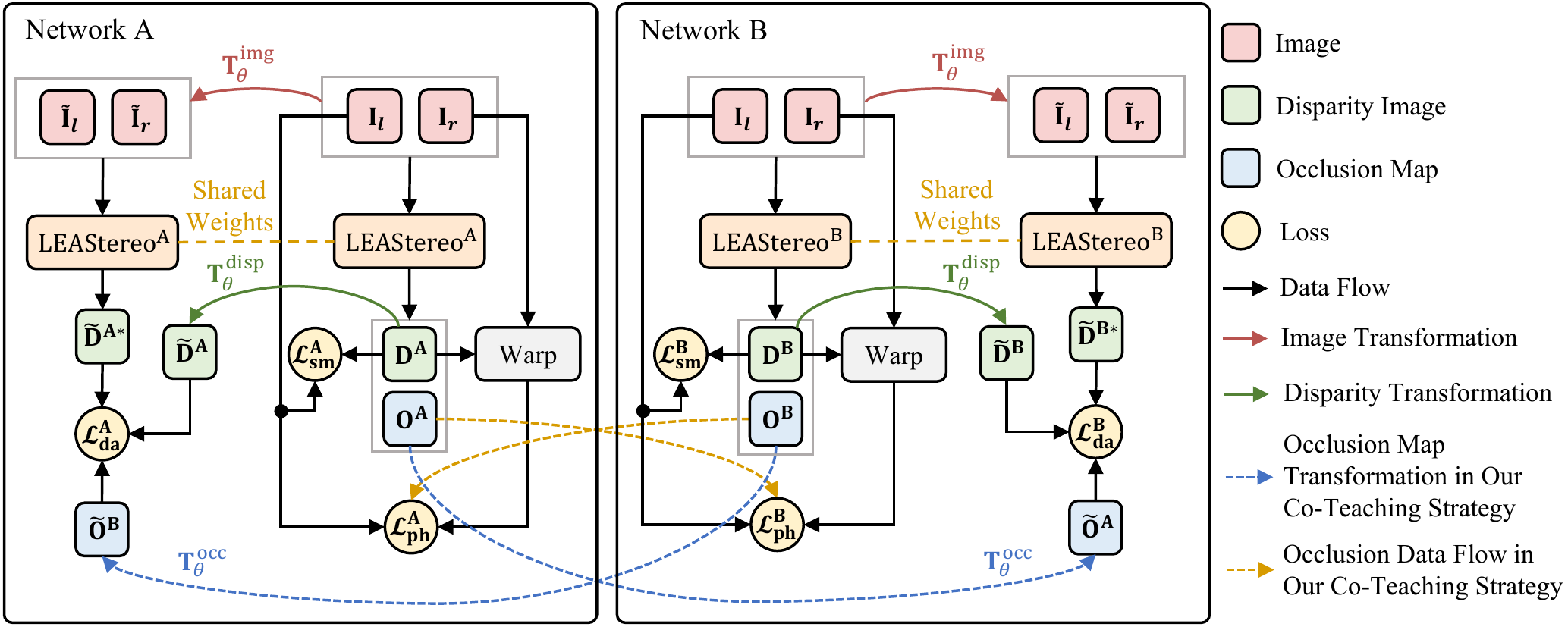}
  \caption{An overview of our CoT-Stereo architecture, where two LEAStereo \cite{cheng2020hierarchical} networks with different initializations teach each other about the occlusions interactively.
  }
  \label{fig.framework}
\end{figure*}

With the limitation of the supervised approaches in mind, many researchers \cite{zhong2017self,tulyakov2017weakly,li2018occlusion,flow2stereo} have resorted to unsupervised techniques, which do not require disparity ground truth to realize stereo matching. These approaches generally train networks by minimizing a hybrid loss, \textit{e.g.}, combining a photometric loss and a smoothness loss \cite{zhong2017self,tulyakov2017weakly}. Some approaches also incorporate occlusion reasoning into the training paradigm to further improve the stereo matching performance \cite{li2018occlusion,flow2stereo}. However, such unsupervised approaches still perform unstably in some regions, especially near occlusions, because a single network can be sensitive to outliers when the disparity ground truth is inaccessible.

To address the instability problem, we propose CoT-Stereo, an unsupervised stereo matching approach. It outperforms all other state-of-the-art unsupervised stereo matching approaches in terms of both accuracy and speed on the KITTI Stereo benchmarks \cite{kitti12,kitti15}, as illustrated in Fig.~\ref{fig.time_error}. Our CoT-Stereo employs a co-teaching framework, as shown in Fig.~\ref{fig.framework}, where two networks (LEAStereo \cite{cheng2020hierarchical} is used as the backbone network) with different initializations interactively teach each other about the occlusions. Our previous work has adopted this co-teaching framework for unsupervised optical flow estimation \cite{wang2020cot}, and in this paper, we employ this framework for unsupervised stereo matching. This framework can significantly improve model's robustness against outliers and further enhance the overall performance of unsupervised stereo matching.

\section{Methodology}
\label{sec.mechodology}
\begin{algorithm*}[t]
  \KwIn{$\Omega^{\mathrm{A}}$ and $\Omega^{\mathrm{B}}$, learning rate $\eta$, constant threshold $\tau$, epoch $T_k$ and $T_{\mathrm{max}}$, iteration $N_{\mathrm{max}}$.}
  \KwOut{$\Omega^{\mathrm{A}}$ and $\Omega^{\mathrm{B}}$.}
  \For{$T = 1 \to T_{\mathrm{max}}$}
  {
  \textbf{Shuffle} training set $\mathcal{D}$\\
  \For{$N = 1 \to N_{\mathrm{max}}$}
  {
  \textbf{Forward} individually to get $\mathbf{D}^{i}$, $\mathbf{O}^{i}$, $\widetilde{\mathbf{D}}^{i}$, $\widetilde{\mathbf{D}}^{i*}$ and $\widetilde{\mathbf{O}}^{i}$, $i \in \{\mathrm{A},\mathrm{B}\}$ \\
  
  \textbf{Set} $\mathbf{O}^{i} \left( \mathbf{O}^{i} > \mathcal{R}(T) \right) = 1$, $i \in \{\mathrm{A},\mathrm{B}\}$ \hspace{2.50cm} $\triangleright$ Omit the pixels with high probability to be occluded \\
  
  \textbf{Compute} $\mathcal{L}^{\mathrm{A}} = \mathcal{L}_{\mathrm{ph}}^{\mathrm{A}}(\mathbf{I}_{l}, \mathbf{I}_{r}, \mathbf{D}^{\mathrm{A}}, \mathbf{O}^{\mathrm{B}}) + \lambda_{1} \cdot \mathcal{L}_{\mathrm{sm}}^{\mathrm{A}} (\mathbf{I}_{l},\mathbf{D}^{\mathrm{A}}) + \lambda_{2} \cdot \mathcal{L}_{\mathrm{da}}^{\mathrm{A}} (\widetilde{\mathbf{D}}^{\mathrm{A}}, \widetilde{\mathbf{D}}^{\mathrm{A}*},\widetilde{\mathbf{O}}^{\mathrm{B}})$\\
  
  \textbf{Compute} $\mathcal{L}^{\mathrm{B}} = \mathcal{L}_{\mathrm{ph}}^{\mathrm{B}}(\mathbf{I}_{l}, \mathbf{I}_{r}, \mathbf{D}^{\mathrm{B}}, \mathbf{O}^{\mathrm{A}}) + \lambda_{1} \cdot \mathcal{L}_{\mathrm{sm}}^{\mathrm{B}} (\mathbf{I}_{l},\mathbf{D}^{\mathrm{B}}) + \lambda_{2} \cdot \mathcal{L}_{\mathrm{da}}^{\mathrm{B}} (\widetilde{\mathbf{D}}^{\mathrm{B}}, \widetilde{\mathbf{D}}^{\mathrm{B}*},\widetilde{\mathbf{O}}^{\mathrm{A}})$\\
  
  \textbf{Update} $\Omega^{i} = \Omega^{i}-\eta \nabla \mathcal{L}^{i}$, $i \in \{\mathrm{A},\mathrm{B}\}$
  }
  \textbf{Update} $\mathcal{R}(T) = 1 - \tau \cdot \min \left\{\frac{T}{T_k}, 1 \right\}$
  }
  \caption{Co-Teaching Strategy}
  \label{alg.co-teaching}
\end{algorithm*}

\subsection{Preliminaries and Loss Functions}
\label{sec.preliminaries_and_loss_functions}
Given a pair of stereo images $\mathbf{I}_{l}$ and $\mathbf{I}_{r}$, the objective of stereo matching is to produce a dense disparity image $\mathbf{D}$. This can be achieved by an off-the-shelf stereo matching network, \eg, LEAStereo \cite{cheng2020hierarchical}. An occlusion map $\mathbf{O}$ indicating each pixel's probability of belonging to the occluded regions can also be computed using the technique proposed in \cite{wang2018occlusion}. Now the problem becomes how to train the network without direct supervision from the disparity ground truth. Following the paradigm of unsupervised stereo matching, we employ a hybrid loss, which combines (a) a photometric loss $\mathcal{L}_{\mathrm{ph}}$, (b) a smoothness loss $\mathcal{L}_{\mathrm{sm}}$, and (c) a data-augmentation loss $\mathcal{L}_{\mathrm{da}}$, to train our CoT-Stereo, as illustrated in Fig.~\ref{fig.framework}. The photometric loss $\mathcal{L}_{\mathrm{ph}}$ can be formulated as a combination of an SSIM term \cite{wang2004image} and an L1 norm term:
\begin{align}
   & \mathcal{L}_{\mathrm{ph}} (\mathbf{I}_{l}, \mathbf{I}_{r}, \mathbf{D}, \mathbf{O}) = \frac{1}{\mathcal{N}} \sum_{\mathbf{p}} \biggl( \alpha \frac{1-\text{SSIM}\left( \mathbf{I}_{l}(\mathbf{p}), \widehat{\mathbf{I}}_{l}(\mathbf{p}) \right)}{2} \notag \\
   & \hspace{1cm} + (1-\alpha)\left\|\mathbf{I}_{l}(\mathbf{p})-\widehat{\mathbf{I}}_{l}(\mathbf{p}) \right\|_{1} \biggr) \cdot \mathcal{S}\left(\overline{\mathbf{O}}(\mathbf{p})\right),
  \label{eq.ph}
\end{align}
where $\widehat{\mathbf{I}}_{l} = \omega(\mathbf{I}_{r},\mathbf{D})$ denotes the warped image from $\mathbf{I}_{r}$ based on $\mathbf{D}$; $\overline{\mathbf{O}}(\mathbf{p}) = 1 - \mathbf{O}(\mathbf{p})$; $\left\| \cdot \right\|_{1}$ denotes the L1 norm; $\mathcal{S}(\cdot)$ denotes the stop-gradient; and $\mathcal{N} = \sum_{\mathbf{p}} \mathcal{S} \left(\overline{\mathbf{O}}(\mathbf{p})\right)$ is a normalizer. Equation~\eqref{eq.ph} shows that $\mathcal{L}_{\mathrm{ph}}$ is an occlusion-aware loss used to penalize the photometric error. Following \cite{li2018occlusion}, we also adopt a smoothness loss $\mathcal{L}_{\mathrm{sm}}$ to smooth the disparity estimations:
\begin{equation}
  \mathcal{L}_{\mathrm{sm}} (\mathbf{I}_{l},\mathbf{D}) = \frac{1}{N_{\mathbf{p}}} \sum_{\mathbf{p}} \sum_{d \in \{x,y\}} \left|\nabla_{d} \mathbf{D}(\mathbf{p})\right| e^{-\left\|\nabla_{d} \mathbf{I}_{l}(\mathbf{p})\right\|_{1}},
  \label{eq.sm}
\end{equation}
where $N_{\mathbf{p}}$ denotes the number of pixels. Moreover, inspired by \cite{liu2020learning}, we adopt a data-augmentation scheme to enable networks to better handle occlusions. Specifically, we first perform transformations $\mathbf{T}_{\theta}^{\mathrm{img}}$, $\mathbf{T}_{\theta}^{\mathrm{disp}}$ and $\mathbf{T}_{\theta}^{\mathrm{occ}}$ (\eg, spatial, occlusion and appearance transformations \cite{liu2020learning}) on $(\mathbf{I}_{l}, \mathbf{I}_{r})$, $\mathbf{D}$ and $\mathbf{O}$ respectively to obtain the augmented samples $\widetilde{\mathbf{I}}_{l}$, $\widetilde{\mathbf{I}}_{r}$, $\widetilde{\mathbf{D}}$ and $\widetilde{\mathbf{O}}$. Please note that, different from $\mathbf{O}$, a higher value in $\widetilde{\mathbf{O}}$ indicates that the pixel is less likely to be occluded in $\widetilde{\mathbf{D}}$ but more likely to be occluded in $\widetilde{\mathbf{D}}^{*}$. Given $\widetilde{\mathbf{I}}_{l}$ and $\widetilde{\mathbf{I}}_{r}$, we can also use LEAStereo \cite{cheng2020hierarchical} to get a disparity estimation $\widetilde{\mathbf{D}}^{*}$. Our data-augmentation loss $\mathcal{L}_{\mathrm{da}}$ is then defined as follows:
{\small
\begin{align}
  \mathcal{L}_{\mathrm{da}} (\widetilde{\mathbf{D}}, \widetilde{\mathbf{D}}^{*},\widetilde{\mathbf{O}}) & = \frac{\sum_{\mathbf{p}} l\left( \lvert \mathcal{S}\left(\widetilde{\mathbf{D}}(\mathbf{p})\right) - \widetilde{\mathbf{D}}^{*}(\mathbf{p}) \rvert \right) \cdot \mathcal{S}\left(\widetilde{\mathbf{O}}(\mathbf{p})\right)}
  {\sum_{\mathbf{p}} \mathcal{S}\left(\widetilde{\mathbf{O}}(\mathbf{p})\right)},\notag                                                                                                                                                                                                                                                 \\
  l(x)                                                                                                  & =\left\{\begin{array}{ll}
    {x-0.5,}     & {x \geq 1} \\
    {x^{2} / 2,} & {x<1}
  \end{array},\right.
  \label{eq.da}
\end{align}
}
where $l(\cdot)$ denotes the smooth L1 loss.

\subsection{Co-Teaching Strategy}
\label{sec.co-teaching_strategy}
Fig.~\ref{fig.framework} and Algorithm~\ref{alg.co-teaching} present the overview of our introduced co-teaching framework, where we simultaneously train two LEAStereo networks: (a) network A (with parameter $\Omega^{\mathrm{A}}$) and (b) network B (with parameter $\Omega^{\mathrm{B}}$). In each mini-batch, the two networks first forward individually to get several outputs (Line 4). Then, we use a dynamic threshold $\mathcal{R}(T)$ to omit the pixels with high occlusion probability (Line 5). $\mathcal{R}(T)$ is designed based on the network memorization mechanism. Specifically, during training, the networks will first learn stereo matching from clear patterns, and then will be gradually affected by outliers \cite{arpit2017closer}. Therefore, $\mathcal{R}(T)$ is initialized as 1 and it decreases gradually as the epochs increase. This helps the networks avoid memorizing outliers (possible inaccurate occlusion estimations) and further improves the performance of unsupervised stereo matching.

Afterwards, we let the two networks swap their estimated occlusion maps and compute their loss functions (Line 6 and 7). Since different networks can learn different types of occlusion and disparity estimations, swapping the occlusion estimations enables the two networks to adaptively correct the inaccurate occlusion estimations, which can further improve the performance of unsupervised stereo matching. Please note that since deep neural networks are highly non-convex, we use two LEAStereo \cite{cheng2020hierarchical} networks with different initializations in our CoT-Stereo. Finally, we update both the parameters of these two networks as well as the dynamic threshold $\mathcal{R}(T)$ (Line 8 and 10).

\section{Experimental Results}
\label{sec.experimental_results}

\subsection{Datasets and Implementation Details}
\label{sec.datasets_and_implementation_details}
For the implementation, we set $\alpha = 0.85$ in Equation~\eqref{eq.ph}. In addition, we set $T_k = 0.2 \cdot T_{\mathrm{max}}$ and $\tau = 0.7$ in Algorithm~\ref{alg.co-teaching}. Moreover, we adopt the Adam optimizer and use a learning rate $\eta = 10^{-4}$ with an exponential decay scheme. Since the two networks present similar performance after convergence, we simply adopt network A for performance evaluation.

We use three public datasets, (a) the Scene Flow \cite{mayer2016large}, (b) the KITTI Stereo 2012 \cite{kitti12}, and (c) the KITTI Stereo 2015 \cite{kitti15} datasets, to validate the effectiveness of our CoT-Stereo. The Scene Flow dataset \cite{mayer2016large} is collected in three different synthetic scenes, while the two KITTI Stereo datasets \cite{kitti12, kitti15} are collected in real-world driving scenarios and have public benchmarks. Two evaluation metrics, (a) the average end-point error (AEPE) that measures the difference between the disparity estimations and ground-truth labels and (b) the percentage of bad pixels (tolerance: 3 pixels) (F1) \cite{kitti12,kitti15}, are adopted for accuracy comparison.

In our experiments, we first conduct ablation studies on the Scene Flow dataset \cite{mayer2016large} to demonstrate the effectiveness of our adopted loss functions and proposed co-teaching strategy, as illustrated in Section~\ref{sec.ablation_study}. Then, we evaluate our CoT-Stereo on the two KITTI Stereo benchmarks \cite{kitti12, kitti15}, as presented in Section~\ref{sec.evaluations_on_the_public_benchmarks}.

\begin{figure*}[t]
  \centering
  \includegraphics[width=0.99\textwidth]{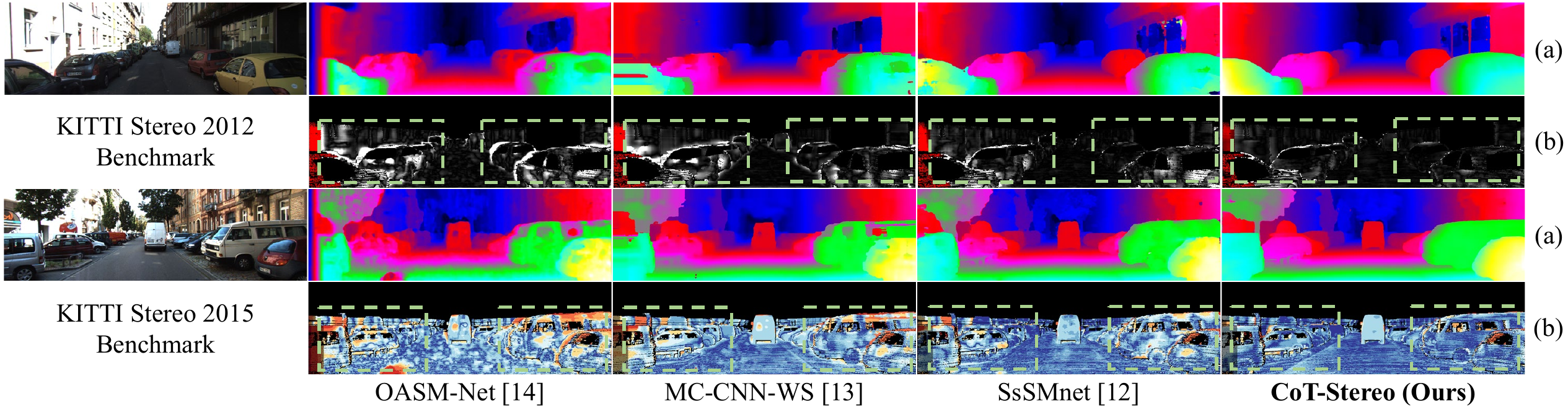}
  \vspace{-0.5em}
  \caption{Examples on the KITTI Stereo benchmarks \cite{kitti12,kitti15}, where rows (a) and (b) show the disparity estimations and the corresponding error maps, respectively. Significantly improved regions are marked with green dashed boxes.}
  \label{fig.benchmark}
  \vspace{-1em}
\end{figure*}

\begin{table}[t]
  \centering
  \caption{Evaluation results of our CoT-Stereo with different setups on the Scene Flow dataset \cite{mayer2016large}. ``Swap'' and ``DT'' denote the occlusion estimation swapping operation and the dynamic threshold selection scheme, respectively. The adopted setup (the best result) is shown in bold type.}
  \begin{tabular}{C{0.6cm}C{0.8cm}C{0.6cm}C{0.6cm}C{0.6cm}C{0.6cm}C{1.6cm}}
    \toprule
    No. & Swap   & DT     & $\mathcal{L}_{\mathrm{ph}}$ & $\mathcal{L}_{\mathrm{sm}}$ & $\mathcal{L}_{\mathrm{da}}$ & AEPE~(px)     \\ \midrule
    (a) & --     & --     & \cmark                      & \cmark                      & \cmark                      & 3.68          \\
    (b) & \cmark & --     & \cmark                      & \cmark                      & \cmark                      & 2.35          \\
    (c) & --     & \cmark & \cmark                      & \cmark                      & \cmark                      & 3.10          \\ \midrule
    (d) & \cmark & \cmark & \cmark                      & --                          & --                          & 4.72          \\
    (e) & \cmark & \cmark & \cmark                      & \cmark                      & --                          & 3.97          \\
    (f) & \cmark & \cmark & \cmark                      & --                          & \cmark                      & 1.86          \\ \midrule
    (g) & \cmark & \cmark & \cmark                      & \cmark                      & \cmark                      & \textbf{1.31} \\
    \bottomrule
  \end{tabular}
  \label{tab.ablation}
  \vspace{-1em}
\end{table}

\begin{table}[t]
  \centering
  \caption{Evaluation results ($\%$) on the KITTI Stereo 2012$^{1}$~\cite{kitti12} and Stereo 2015$^{2}$ \cite{kitti15} benchmarks. ``S'' denotes supervised approaches. ``Noc'' and ``All'' represent the F1 for non-occluded pixels and all pixels, respectively \cite{kitti12,kitti15}. Best results for supervised and unsupervised approaches are both shown in bold type.}
  \begin{tabular}{L{2.8cm}C{0.4cm}C{0.6cm}C{0.6cm}C{0.6cm}C{0.6cm}}
    \toprule
    \multicolumn{1}{l}{\multirow{2}{*}{Approach}} & \multicolumn{1}{l}{\multirow{2}{*}{S}} & \multicolumn{2}{c}{KITTI 2012} & \multicolumn{2}{c}{KITTI 2015}                                 \\ \cmidrule(l){3-4} \cmidrule(l){5-6}
    \multicolumn{1}{c}{}                          & \multicolumn{1}{c}{}                   & Noc                            & All                            & Noc           & All           \\ \midrule
    PSMNet \cite{psmnet}                          & \cmark                                 & 1.49                           & 1.89                           & 2.14          & 2.32          \\
    GwcNet-gc \cite{gwcnet}                       & \cmark                                 & 1.32                           & 1.70                           & 1.92          & 2.11          \\
    LEAStereo \cite{cheng2020hierarchical}        & \cmark                                 & \textbf{1.13}                  & \textbf{1.45}                  & \textbf{1.51} & \textbf{1.65} \\ \midrule
    OASM-Net \cite{li2018occlusion}               & --                                     & 6.39                           & 8.60                           & 7.39          & 8.98          \\
    Flow2Stereo \cite{flow2stereo}                & --                                     & 4.58                           & 5.11                           & 6.29          & 6.61          \\
    MC-CNN-WS \cite{tulyakov2017weakly}           & --                                     & 3.02                           & 4.45                           & 4.11          & 4.97          \\
    SsSMnet \cite{zhong2017self}                  & --                                     & 2.30                           & 3.00                           & 3.06          & 3.40          \\
    \textbf{CoT-Stereo (Ours)}                    & --                                     & \textbf{1.82}                  & \textbf{2.32}                  & \textbf{2.43} & \textbf{2.68} \\ \bottomrule
  \end{tabular}
  \label{tab.disparity}
\end{table}

\subsection{Ablation Study}
\label{sec.ablation_study}
Table~\ref{tab.ablation} presents the evaluation results of our CoT-Stereo with different setups on the Scene Flow dataset \cite{mayer2016large}. For our proposed co-teaching strategy, (a)--(c) and (g) of Table~\ref{tab.ablation} demonstrate the effectiveness of the occlusion estimation swapping operation and the dynamic threshold selection scheme, which can effectively improve unsupervised stereo matching. Additionally, we can clearly observe that the combination of the three loss functions can effectively improve the performance, as shown in (d)--(g) of Table~\ref{tab.ablation}. Moreover, (g) in Table~\ref{tab.ablation} denotes the adopted setup, which validates the effectiveness of our adopted loss functions and proposed co-teaching strategy.

\subsection{Evaluations on the Public Benchmarks}
\label{sec.evaluations_on_the_public_benchmarks}
Table~\ref{tab.disparity} shows the online leaderboards of the KITTI Stereo 2012 \cite{kitti12} and Stereo 2015 \cite{kitti15} benchmarks, and Fig.~\ref{fig.time_error} visualizes the results on the KITTI Stereo 2015 benchmark. We can observe that our CoT-Stereo outperforms all other state-of-the-art unsupervised stereo matching approaches in terms of both accuracy and speed, which demonstrates the effectiveness of the occlusion estimation swapping operation and the dynamic threshold selection scheme for unsupervised stereo matching. Excitingly, our CoT-Stereo can even present competitive performance compared with the state-of-the-art supervised approaches. Examples on the KITTI Stereo benchmarks are shown in Fig.~\ref{fig.benchmark}, where it is evident that our CoT-Stereo can generate more robust and accurate disparity estimations. All the analysis proves the excellent performance of our CoT-Stereo for unsupervised stereo matching.

\section{Conclusion and Future Work}
\label{sec.conclusion_and_future_work}
This paper proposed a novel co-teaching strategy for unsupervised stereo matching, which consists of a dynamic threshold selection scheme and an occlusion estimation swapping operation. The former ensures that the networks do not memorize possible outliers, while the latter enables the two networks to adaptively correct the inaccurate occlusion estimations and further improve the performance of unsupervised stereo matching. Extensive experimental results on the KITTI Stereo benchmarks showed that our approach, CoT-Stereo, outperforms all other state-of-the-art unsupervised stereo matching approaches in terms of both accuracy and speed.

\footnotetext[1]{\url{http://www.cvlibs.net/datasets/kitti/eval_stereo_flow.php?benchmark=stereo}}
\footnotetext[2]{\url{http://www.cvlibs.net/datasets/kitti/eval_scene_flow.php?benchmark=stereo}}


\clearpage
\bibliographystyle{IEEEbib}
\bibliography{egbib}

\end{document}